\documentclass[5p, a4paper]{elsarticle}

\usepackage{amssymb}
\usepackage{graphicx}
\usepackage{subfigure}
\usepackage{amsmath}  

\usepackage{algorithm}
\usepackage{algpseudocode}

\usepackage{float}
\restylefloat{table}

\let\today\relax
\makeatletter
\def\ps@pprintTitle{%
    \let\@oddhead\@empty
    \let\@evenhead\@empty
    \def\@oddfoot{\footnotesize\itshape
         {Submitted preprint} \hfill\today}%
    \let\@evenfoot\@oddfoot
    }
\makeatother

\begin{document}

\begin{frontmatter}



\title{MultiVehicle Simulator (MVSim): lightweight dynamics simulator for multiagents and mobile robotics research}



\author{José-Luis Blanco-Claraco\corref{cor1}\fnref{label1}}
\ead{jlblanco@ual.es}

\author{Borys Tymchenko\fnref{label2}}


\author{Francisco José Mañas-Alvarez\fnref{label3}}

\author{Fernando Cañadas-Aránega\fnref{label4}}
\author{Ángel López-Gázquez\fnref{label4}}
\author{José Carlos Moreno\fnref{label4}}

\fntext[label1]{Engineering Department, University of Almer\'{\i}a, 
Agrifood Campus of International Excellence (ceiA3), and CIESOL, 04120 Almer\'{\i}a, Spain}
\fntext[label2]{Deci AI, Tel Aviv (Israel)}
\fntext[label3]{Department of Computer Sciences and Automatic Control, Universidad Nacional de Educaci\'{\o}n a Distancia (UNED), Madrid (Spain)}
\cortext[cor1]{Corresponding author}
\fntext[label4]{Department of Informatics, University of Almer\'{\i}a, 
Agrifood Campus of International Excellence (ceiA3), and CIESOL, 04120 Almer\'{\i}a, Spain}


\begin{abstract}
Development of applications related to closed-loop control requires
either testing on the field or on a realistic simulator, with the latter being
more convenient, inexpensive, safe, and leading to shorter development cycles.
To address that need, the present work introduces MVSim, a simulator for multiple vehicles or robots capable 
of running dozens of agents in simple scenarios, or a handful of them in 
complex scenarios. 
MVSim employs realistic physics-grounded friction models for tire-ground interaction, 
and aims at 
accurate and GPU-accelerated simulation of most common modern sensors employed in mobile robotics and autonomous vehicle research, such as depth and RGB cameras, or 2D and 3D LiDAR scanners. All depth-related sensors are able to accurately measure distances to
3D models provided by the user to define custom world elements.
Efficient simulation is achieved by means of focusing on ground vehicles,
which allows the use of a simplified 2D physics engine for body collisions
while solving wheel-ground interaction forces separately.
The core parts of the system are written in C++ for maximum efficiency, 
while Python, ROS 1, and ROS 2 wrappers are also offered for easy integration
into user systems. A custom publish/subscribe protocol based on ZeroMQ (ZMQ) is 
defined to allow for multiprocess applications to access or modify a running
simulation.
This simulator enables and makes easier to do research and development 
on vehicular dynamics, 
autonomous navigation algorithms, and simultaneous localization and mapping (SLAM) methods.
\end{abstract}


\end{frontmatter}

\section{Motivation and significance}
\label{sect:motiv}


Research and development in mobile robotics or autonomous vehicles
have final goals related to physical prototypes, which leads to expensive
and laborious field experiments to tune and debug new implementations.
In the industry, this turns out into long development cycles, increasing the time-to-market of new products or services.
\begin{figure}[h]
	\centering
	\includegraphics[width=0.98\columnwidth]{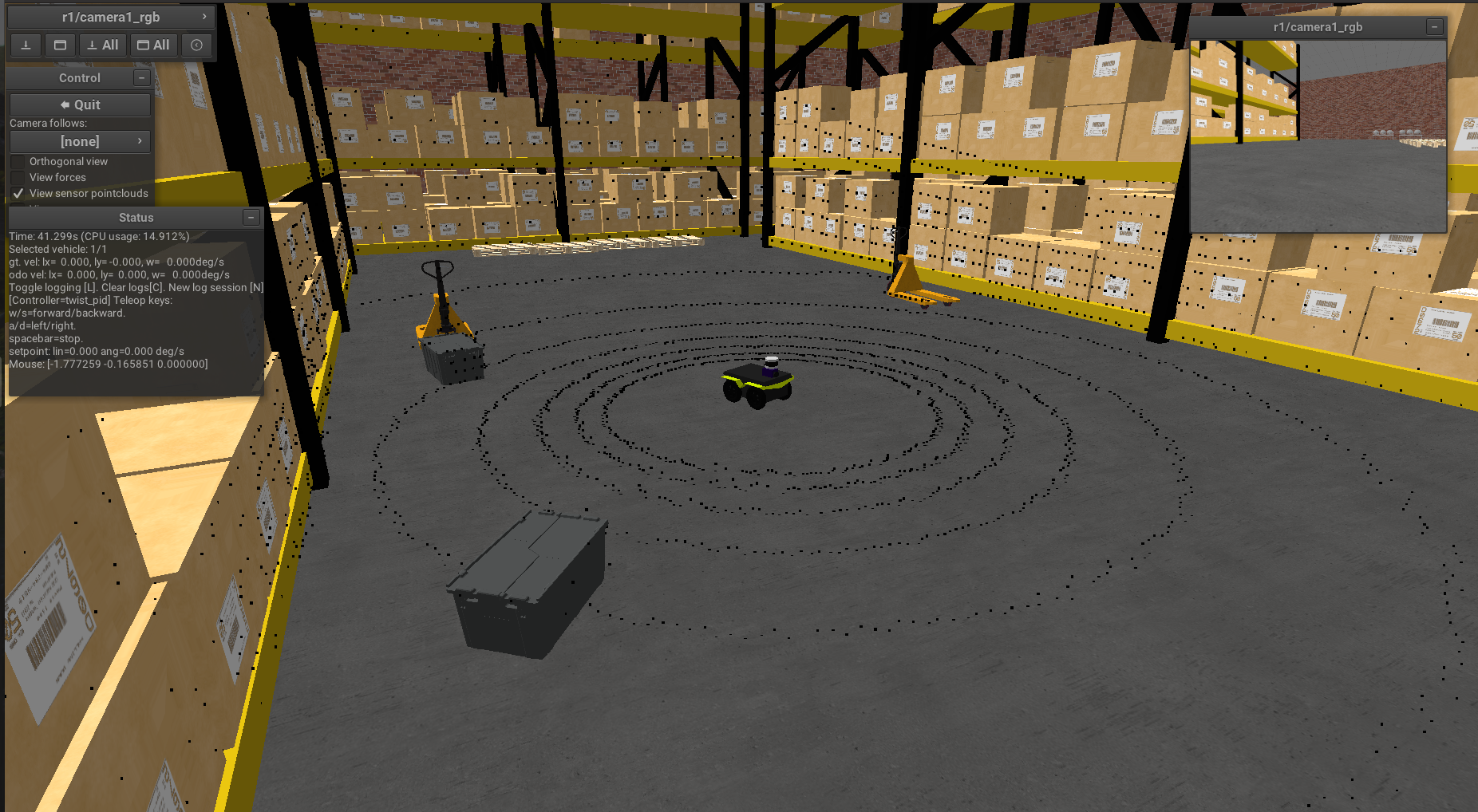}
	\caption{Screenshot of the simulator running the warehouse example world. Refer to discussion in section~\ref{sect:examples}}.
	\label{fig:screenshot.warehouse}
\end{figure}
%
In particular, the present work has a great potential in addressing three scientific problems:
(i) simultaneous localization and mapping (SLAM), (ii) autonomous navigation, 
and (iii) steering control.
SLAM is the problem of a mobile agent equipped with exteroceptive sensors to incrementally build a map of its 
environment as it moves around, autonomously or teleoperated \cite{grisetti2010tutorial, cadena2016past}.
Autonomous navigation is the problem of safely taking a vehicle or robot
from one initial pose to a goal one. 
Typically, this large research field distinguishes between 
global path planners (find a path given a map) and
local planners (trying to follow a predefined path, avoiding unexpected and dynamic obstacles) \cite{kavraki2016motion}.
Finally, steering control and modeling of ground vehicles \cite{mandow2007experimental}
address the issue of finding accurate mathematical models for the dynamics and kinematics of steering 
in such vehicles, and all related applications as state-observers \cite{reina2017vehicle} 
or control \cite{wang2016robust}.


\begin{figure*}
\centering
\subfigure[Module: mvsim-comms]{\includegraphics[height=0.2\textheight]{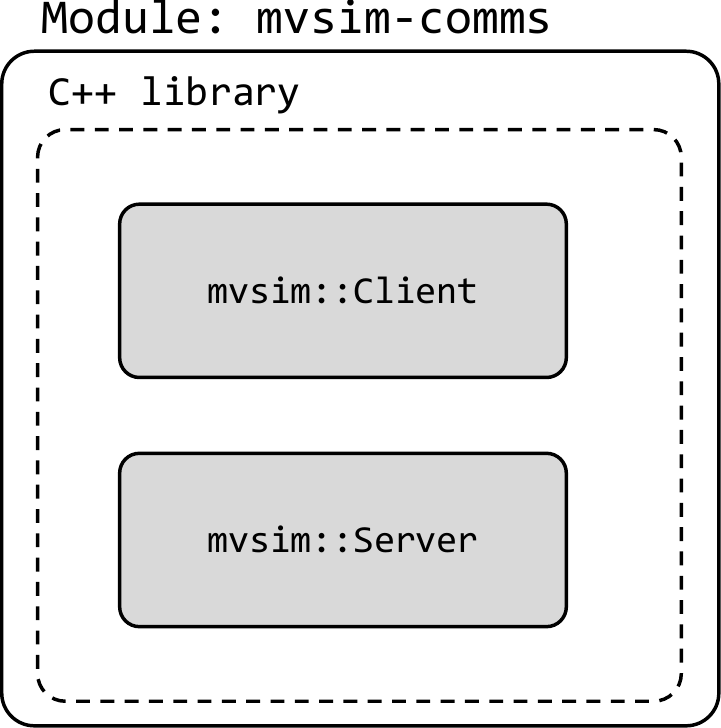}}
\hspace{5mm}
\subfigure[Module: mvsim-msgs]{\includegraphics[height=0.2\textheight]{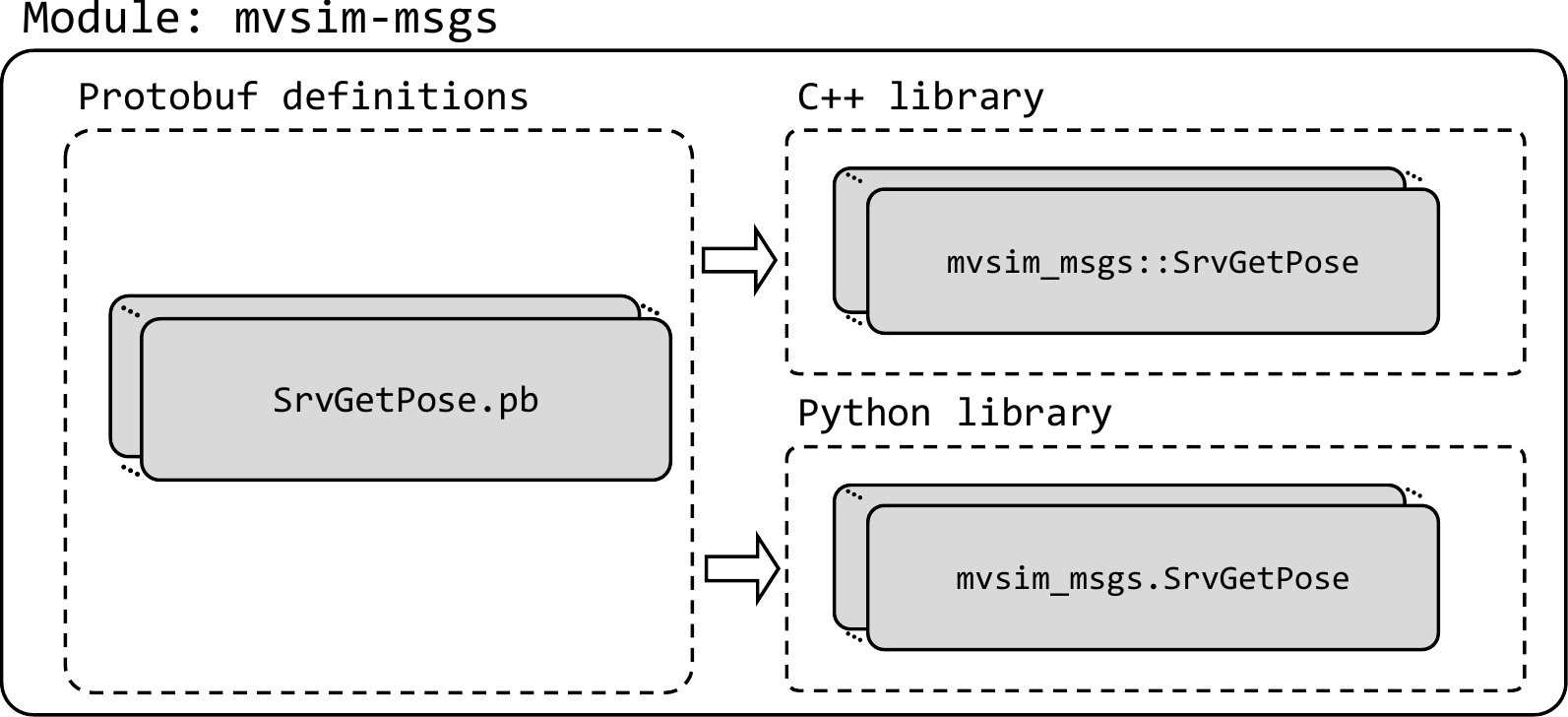}}
~
\subfigure[Module: mvsim-simulator]{\includegraphics[width=0.8\textwidth]{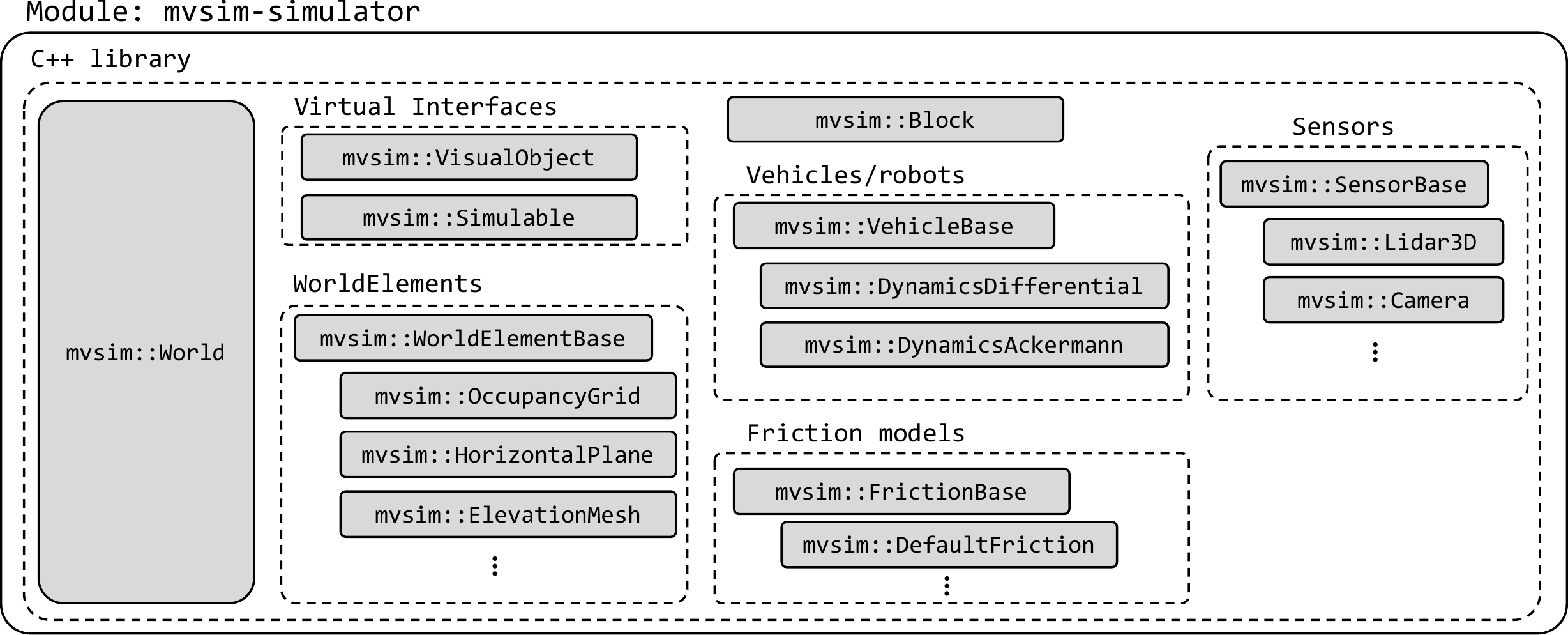}}
\caption{Overview of the modules provided by MVSIM: 
(a) The communications (messages transport) module,
(b) the messages definitions module,
and (c) the simulation module.}
\label{fig:arch.modules}
\end{figure*}

\begin{figure*}
\centering
\includegraphics[width=0.8\textwidth]{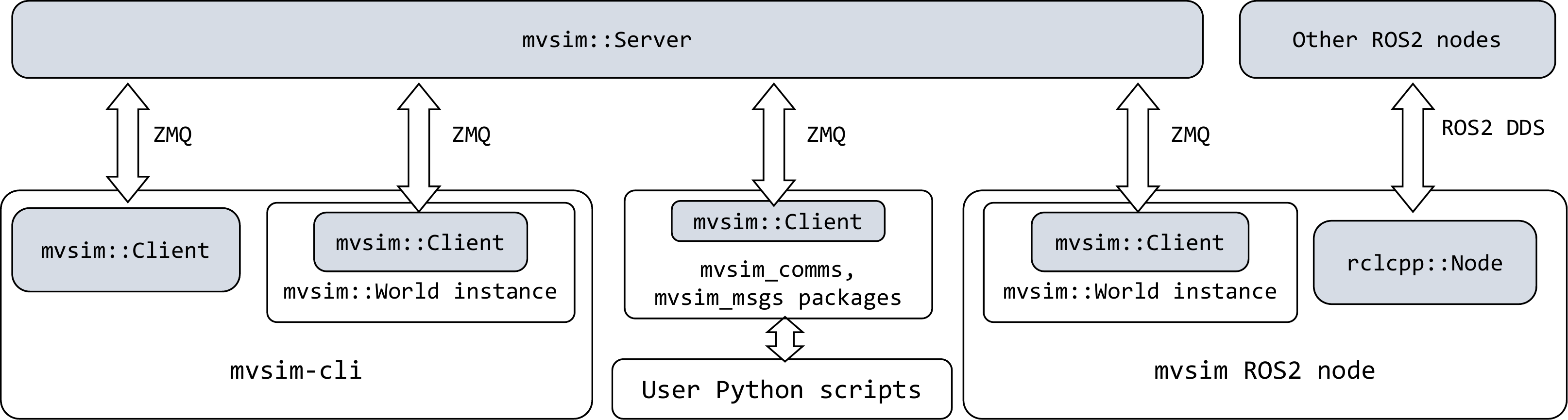}
\caption{High-level diagram of potential connections 
between different processes using MVSIM.}
\label{fig:arch.programs}
\end{figure*}


Users can use MVSim in all major operating systems (GNU/Linux, OSX, Windows).
The standalone executable \texttt{mvsim-cli} is provided 
as a central hub to most common operations, such as launching a simulation, 
or listing and inspecting published topics.
A python library is also provided to allow users to easily communicate with
a running simulation, from the same or a different machine.
Finally, a node for ROS 1 and ROS 2 is provided to act as an intermediary 
between an MVSim simulation and other ROS nodes, using standardized APIs
and conventions as \texttt{tf} frames \cite{meeussen2010rep} or 
sensor message types \cite{rico2022concise}.


There exist other software projects with similarities to MVSim.
The most notorious are: Player/Stage, Gazebo, and Webots.
Player/Stage \cite{gerkey2003player} is an open-source project consisting of a set of software tools used for multi-robot systems (MRS) control. Two main parts are considered: Player and Stage. Player is a device server providing a transparent access to the robot, and Stage is the simulator for moving and sensing multiple robots in a 2D indoor environment.
Gazebo \cite{koenig2004design}, specially developed to take complex outdoor environments into account, is an open-source and freely available MRS simulation backend as Stage, also compatible with Player, but operating in a 3D world and simulating physical interactions between objects. Gazebo also includes its own native interface with the robots. 
However, it lacks flexibility in the selection of 
friction models between vehicle tires and the ground,
which may be specially important depending on the type of
robot and world simulated.
Webots \cite{Michel2004}, as Gazebo, is an open-source robot simulator capable of working with all kind of robots (UGV, automobiles, UAV, bipeds, manipulators...), and MRS, in outdoor environments, and incorporating physics for each component of the world.


\section{Software description}
\label{}


Next, the overall architecture of the software is exposed in section~\ref{sect:arch} together with some implementation details
regarding the physics simulation. Later on, section~\ref{sect:func} 
provides a summary of the functionality offered by the proposal.

\subsection{Software Architecture}
\label{sect:arch}


The project comprises libraries and executables, 
with the formers (C++ and Pyhton libraries)
split into three modules
as shown in Figure~\ref{fig:arch.modules}.
The \texttt{mvsim-comms} module is responsible of providing
top-level client-server functionality for publish-subscribe
and remote service invocation. Data transport is implemented
by means of ZeroMQ sockets for their
portability, robustness, and efficiency \cite{hintjens2013zeromq}.
Data interchange is done by means of messages, defined 
in the \texttt{mvsim-msgs} module by means of an 
interface definition language (IDL), in particular,
Google Protobuf. IDL definitions are compiled into
a C++ library and a Python library.
Finally, the \texttt{mvsim-simulator} module
holds the main functionalities of the software: 
loading and parsing world definition files,
running simulations,
exposing a graphical interface for the user to 
manipulate and explore running simulations, 
and
simulating and publishing sensors.

These libraries can then be used in different ways, 
as illustrated in Figure~\ref{fig:arch.programs}.
Firstly, the \texttt{mvsim-cli} program is provided as a
central hub to most non programming-related functionality, 
such as: launching a simulation for a particular world 
definition file and exploring it via a GUI, listing and getting details on connected communication clients and advertised topics, 
or analyzing the rate at which a topic is being published.
Note that invoking the \texttt{"mvsim-cli launch"} subcommand
spans a communication server,
to which the client within \texttt{mvsim::World} connects to.
Another way to run a simulation (and spanning a server) is 
launching the provided ROS node. Note that, in that case,
the MVSim simulation is accessible simultaneously via
our custom Python library, and the standard ROS Pub/Sub mechanism.
It is worth mentioning that headless execution of the simulation is also possible (including GPU-accelerated sensor simulation), 
enabling the usage of MVSim within containerized
Continuous Integration (CI) pipelines.

Regarding physics simulation, efficiency is achieved by means of a 
hierarchical approach: on a first (higher) level, rigid objects 
(named "blocks" in MVSim and which may represent fixed or mobile obstacles) and vehicles (which include wheels as their unique mobile parts) are modeled as 2D entities using the Box2D C++ library \cite{catto2005iterative}.
For vehicles or objects with custom 3D meshes, MVSim evaluates their equivalent 2D collision planar polygon.
The Box2D engine then provides collision detection and runs physics simulation on the basis
of forces and torques applied to each rigid body, whose calculation is described below.
Simulation runs forward in time by applying numerical integration on the state vector of all mobile objects with a fixed time step. Before and after running each such step,
the Algorithms~\ref{alg:pre} and ~\ref{alg:post} are
applied to account for the second (lower) layer of 
the hierarchical physics simulation.
Next we provide details on the most relevant parts of these algorithms to explain the underlying mechanical model.
Kinematics (Ackermann vs. differential-driven vehicles) and 
PID controllers to achieve closed-loop control of wheels velocity
are already handled in \texttt{invoke\_motor\_controllers()}
in Algorithm~\ref{alg:pre}. Note that closed-loop velocity control 
is achieved taking simulated odometry readings as the controller input, 
thus wheels slippage will lead to wrong odometry estimations,
accurately replicating the consequences of slippage on real vehicles.

Each vehicle or robot ($V$)
is modeled
as a rigid body ("chassis") with a given mass and inertia tensor, whose SE(3) pose \cite{sola2018micro,blanco2021tutorial} 
with respect to a global world origin $O$
can be defined as an homogeneous transformation matrix $^O \mathbf{T}_V$, comprising a translation vector $^O\mathbf{t}_V$
and a $3\times 3$ rotation matrix $^O\mathbf{R}_V$.
Note the use of 3D coordinates and orientation, SE(3), instead of the
equivalent planar SE(2) poses, despite the use of an underlying 2D physics engine. That is because MVSim can provide certain degrees of freedom for vehicles to move up and down and tilt as they transverse digital terrain elevation models, for example.

\begin{figure}[t]
	\centering
	\subfigure[]{\includegraphics[width=0.5\columnwidth]{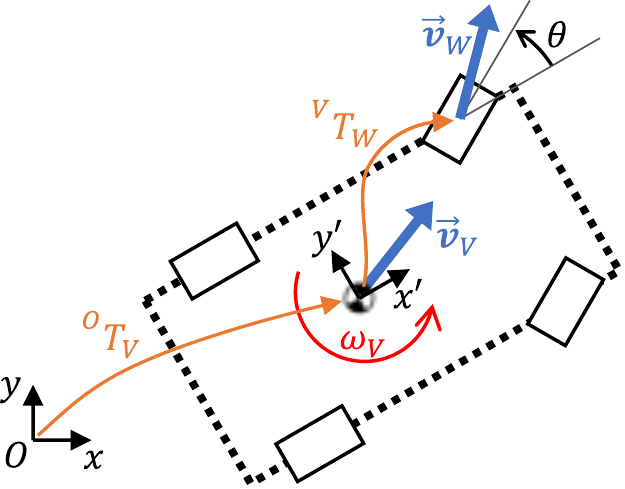}}
	\subfigure[]{\includegraphics[width=0.85\columnwidth]{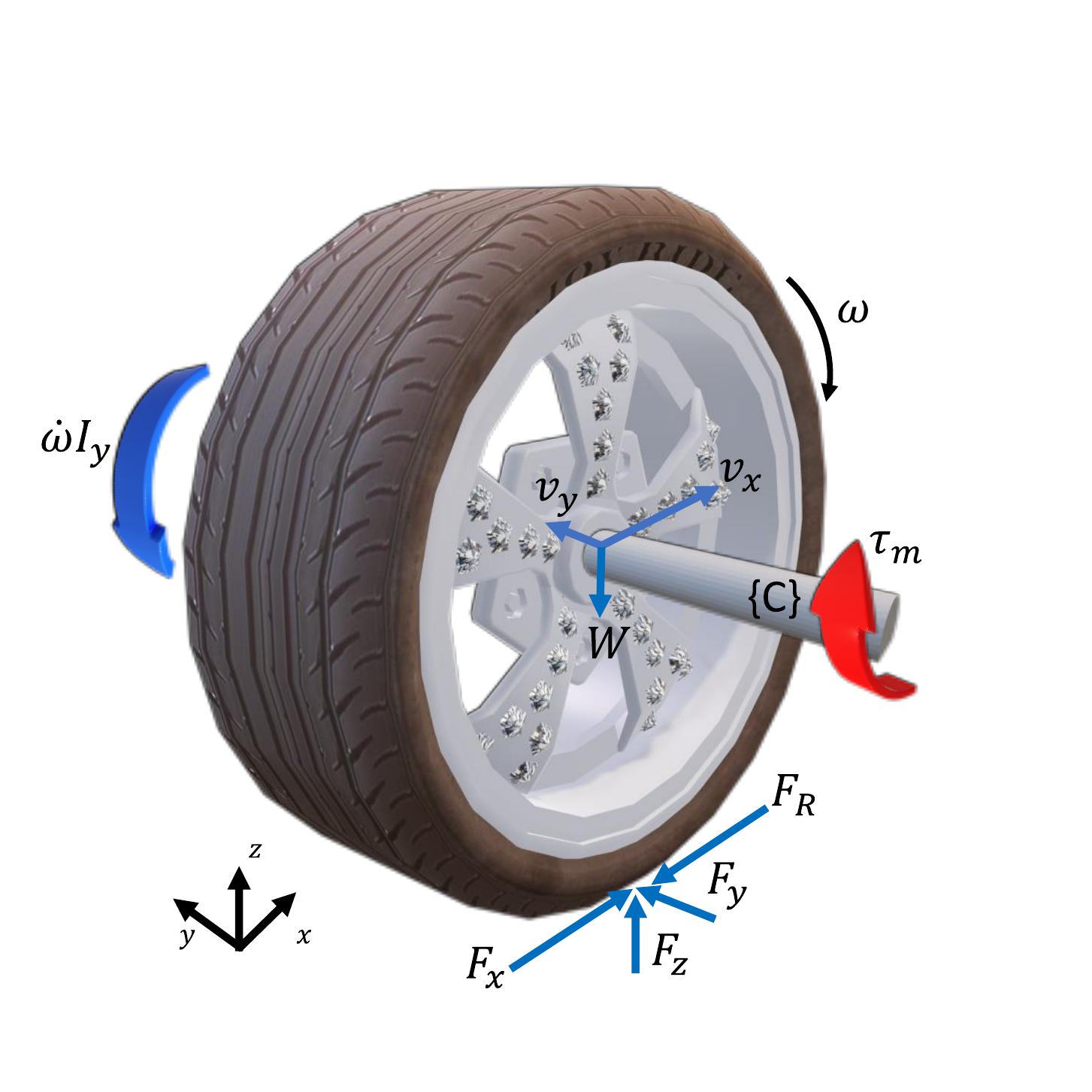}}	
	\caption{(a) Kinematics model of a wheel $W$ on a vehicle $V$. (b) Free body diagram of the mechanical problem involved in determining reaction forces at the wheel-ground contact point. Here, $x$ is always the wheel local forward-facing direction and $z$ points up, $\dot{\omega}$ is the instantaneous angular acceleration of the wheel spinning motion, and $I_y$ the inertia tensor component related to axial ($y$) rotation.}.
	\label{fig:physics}
\end{figure}

Let $\mathbf{v}_V$ be the velocity vector of the vehicle $V$ reference point (which may or may not coincide with the center of mass), and 
$\boldsymbol{\omega}_V$ the vehicle angular velocity, 
as illustrated in Figure~\ref{fig:physics}(a).
Then, rigid body kinematics leads to the following expression 
for the velocity vector of a wheel ($W$) center point, expressed in global coordinates ($O$):

\begin{equation}
\label{eq:vW_vV}
\mathbf{v}_W = \mathbf{v}_V + \boldsymbol{\omega}_V \times {}^{V}\mathbf{t}_W
\end{equation}

\noindent with $\times$ the cross product and ${}^{V}\mathbf{t}_W$ the translational part of the relative pose ${}^{V}T_W$ of the wheel $W$ with respect to the vehicle reference frame $V$. For planar motion, the cross product has the optimized form shown in line 8 of 
Algorithm~\ref{alg:pre}.
To evaluate the tire-ground friction model the relevant 
components of the velocity vector of the wheel center point
are those in local coordinates with respect to the wheel, 
i.e. with the local $x$ axis always pointing forward along the
longitudinal wheel axis. 
Let $\theta$ be the rotation of the wheel with respect to the
vehicle longitudinal axis, as in Figure~\ref{fig:physics}(a).
Then, the velocity vector in the wheel frame $W$ is related
to the velocity in global coordinates from Eq.~(\ref{eq:vW_vV})
by 
$^W{\mathbf{v}}_W = \mathbf{R}_z(\theta)^{-1} \mathbf{v}_W$
(line 9 of 
Algorithm~\ref{alg:pre})
with $\mathbf{R}_z(\cdot)$ the rotation matrix for rotations around the $z$ axis.

Once we know the chassis weight $W$ loaded on each wheel axis,
the motor torque $\tau_m$ exerted at each such axis, and the
instantaneous local velocity vector $(v_x,v_y)$ of the wheel center point, 
we can apply a friction model to determine the resulting longitudinal ($F_x$) and lateral ($F_y$) forces that the wheel is exerting on 
the vehicle chassis, effectively driving the vehicle motion.
Different friction models are implemented in MVSim, 
but at present all of them rely on
the mechanical model depicted in Figure~\ref{fig:physics}(b).
Note that $C$ denotes here an optional damping factor, modeling velocity-proportional losses in the applied torque, $\omega$ 
is the instantaneous wheel spinning angular velocity, 
and $F_R$ is the longitudinal friction force (essential to prevent wheel slippage). In friction-related mechanical problems
we have upper bounds for the absolute value of friction forces,
but determining whether those limits are hit or not requires two-steps:
(i) to solve equilibrium equations assuming friction is large enough
to fulfill the no-slippage condition, then (ii) clamp the friction force if needed and solve again if slippage do happen.
Note that static equilibrium equations can be split into three separable axis.
Firstly, vertical forces equilibrium reveals that $F_z=W$, with $W$ here the sum of the partial chassis weight and the wheel weight itself.
Therefore, the maximum friction force using a simple 
dry-friction model with friction coefficient $\mu$
is $F_{max} = \mu W$.
Secondly, lateral equilibrium requires the determination of the 
friction force $F_y$ required to avoid lateral slippage induced
by the velocity component $v_y$: the lateral acceleration 
required to reduce $v_y$ to zero for a simulation time step $dt$
is $a_y = - v_y / dt$, thus given Newton's second law of motion
the required force is $F_y = -\frac{v_y}{dt}\frac{W}{g}$ with $g$ the gravity acceleration. This force is clamped to the range $[-F_{max},F_{max}]$, leading to lateral wheel slippage when the bounds are hit.
Thirdly, equilibrium of longitudinal forces and moment on the local axis $y$ finally gives us the value for $F_x$.
In this case, the kinematic constraint (rolling without sliding)
would imply changing the wheel angular speed from $w$ to 
$\omega^\star = v_x / R$ with $R$ the wheel effective radius, 
hence imposing a desired angular acceleration of
$\dot{\omega}^\star=(\omega^\star-\omega)/dt$.
Taking moments from the wheel axis in Figure~\ref{fig:physics}(b)
leads to the equilibrium equation:

\begin{equation}
\label{eq:equil.wheel}
\tau_m - \dot{\omega} I_y + R F_R - R F_x = 0
\end{equation}

\noindent where $F_R$ is an optional rolling friction
(e.g. MVSim implements 
the realistic Ward-Iagnemma model in \cite{ward2008dynamic}).
Once the friction force $F_x$ is determined, if saturation
happens due to the existence of the upper bound $F_{max}$, 
Eq.~(\ref{eq:equil.wheel}) is evaluated again with $F_x=\pm F_{max}$ 
to determine the actual spinning angular acceleration
$\dot{\omega}$ given that longitudinal slippage was unavoidable.

\begin{algorithm}[t]
\caption{\texttt{pre\_timestep()} algorithm}\label{alg:pre}
\begin{algorithmic}[1]
\ForAll{v in vehicles}
\State \# Evaluate chassis weight $\mathbf{W}$ on each wheel:
\State $\mathbf{W}[]\quad \gets$ v.weight\_on\_wheels()
\State \# Evaluate motor torques $\mathbf{T}$:
\State $\mathbf{T}[]\quad \gets$ v.invoke\_motor\_controllers()
\ForAll{i,w in enumerate(v.wheels)}
\State \# Wheel center point velocity vector:
\State $\mathbf{v}_w \gets \mathbf{v}_g + ( -\omega*w.y, \omega*w.x )$ \Comment{See~Eq.~(\ref{eq:vW_vV})}
\State $\mathbf{v}_{local} \gets \text{rotate\_z}(\mathbf{v}_w, \theta_i)$
\State $\{F_x,F_y\} \gets v.friction\_model(\mathbf{v}_{local}, \mathbf{T}[i],\mathbf{W}[i])$
\State \# Apply force to vehicle chassis at wheel center:
\State v.apply\_force($\{F_x,F_y\}$, $\mathbf{t}_W$) %
\EndFor
\EndFor
\end{algorithmic}
\end{algorithm}

\begin{algorithm}[t]
\caption{\texttt{post\_timestep()} algorithm}\label{alg:post}
\begin{algorithmic}[1]
\Require dt = simulation time step
\ForAll{v in vehicles}
\ForAll{s in v.sensors} \# Process all sensors
\State s.process()
\EndFor
\State \# Integrate spinning of wheels:
\ForAll{w in v.wheels}
\State  $w.phi \gets wrap\_pi(w.phi + w.omega * dt)$
\EndFor
\EndFor
\end{algorithmic}
\end{algorithm}

Finally, it is worth mentioning that MVSim supports
vehicles or robots with 3 or more wheels to move on
terrain models defined by elevation maps. This is 
accomplished by determining the elevation of each
wheel-ground contact point with respect to the
reference level, then applying the 
quaternion Horn method \cite{horn1987closed} 
to solve for the optimal vehicle attitude.

\subsection{Software Functionalities}
\label{sect:func}

MVSim allows running simulations of one or more vehicles
observing collisions, realistic rigid body kinematics 
and wheel-ground friction models, as well as sensor simulation.
Sensor data is accessible via either our custom ZMQ-based pub/sub
communication system (C++ and Python interfaces)
or via ROS1 or ROS2 standard sensor messages.
Simulations can run in real wall-clock time or 
at a different rate. 
For example, faster than real-time runs can be used in headless instances to support simulation-based unit tests of mobile robotic software.
Implemented sensors at present include pin-hole cameras, 
RGBD (depth) cameras, 2D and 3D LiDARs. In all these cases, 
sensor simulation exploits GPU acceleration via an OpenGL pipeline to
achieve generation of millions of 3D LiDAR points per second in real-time.
Each vehicle is equipped with a configurable PID controller to
generate the motor torques required to accomplish the desired 
vehicle linear and angular speed set-points.
Additionally, vehicles have an optional high-speed logging 
system to log all the internal physics details (frictions, forces, torques, etc.) to CSV files for posterior debug and analysis.
World models are defined by means of XML files whose format is 
detailed in the on-line project documentation page. 
There is support for including other XML files within a given 
file, parsing and replacement of environment variables, 
mathematical expression evaluation, and flow control-like 
structures such as \texttt{for} loops at the XML document level.


\begin{figure}[t]
	\centering
	\subfigure{ \includegraphics[width=0.98\columnwidth]{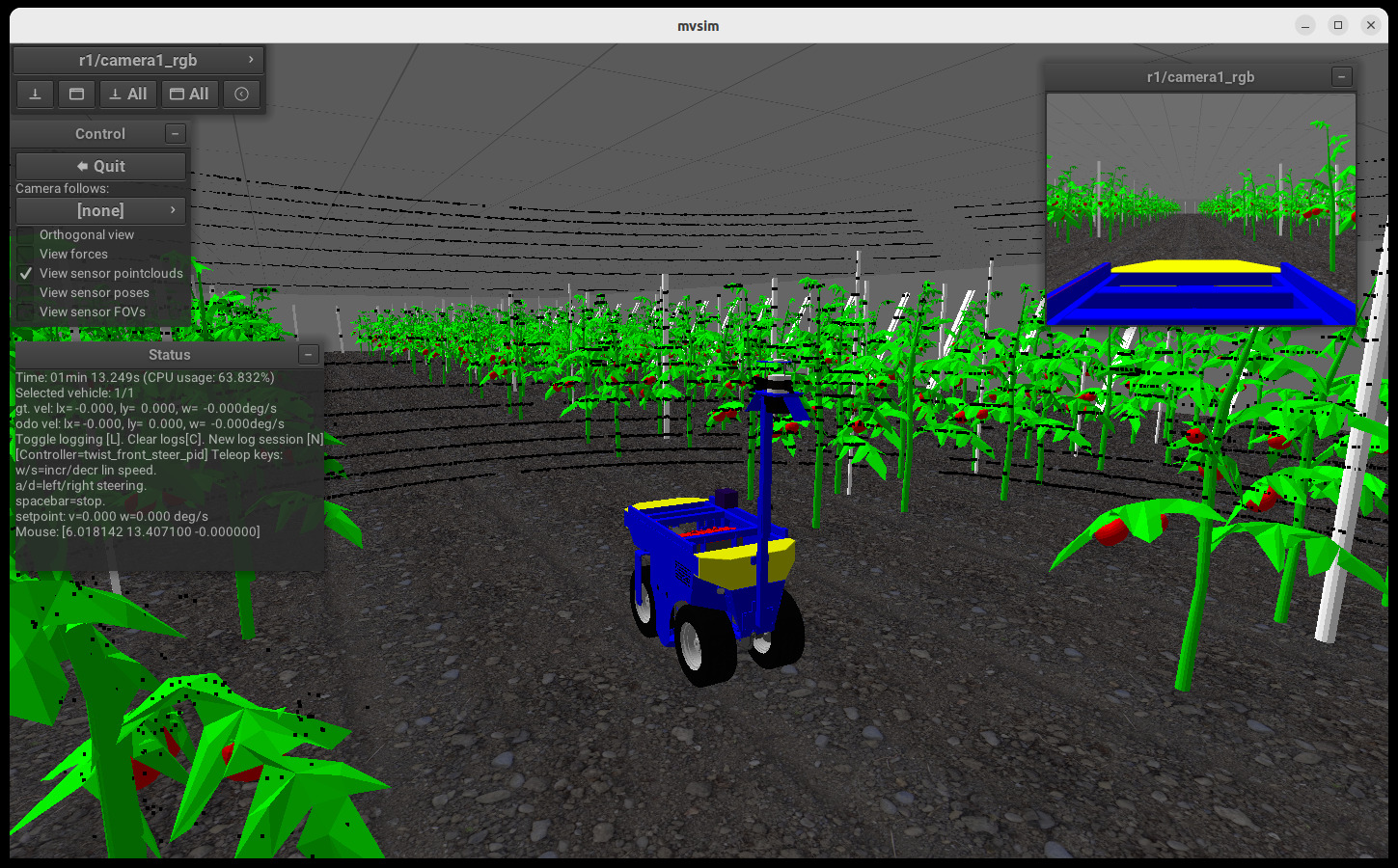}}
	\subfigure{ \includegraphics[width=0.98\columnwidth]{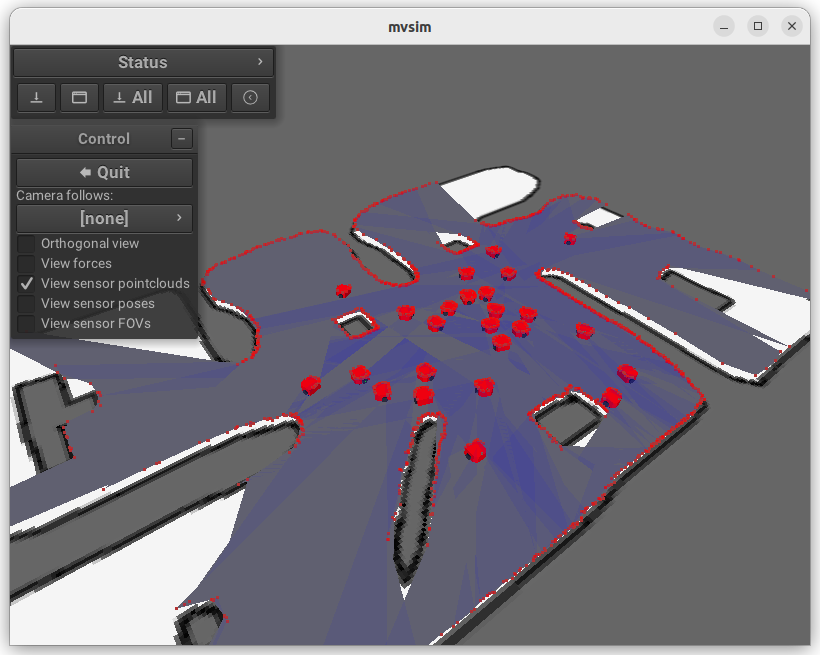}}
	\caption{Screenshots of demo world files provided with the package. See discussion of section~\ref{sect:examples}.}.
	\label{fig:screenshot.demos}
\end{figure}

\section{Illustrative Examples}
\label{sect:examples}


The project repository contains several example files
illustrating the different possibilities of the simulator 
and the XML-definition language under the 
\texttt{mvsim\_tutorials} subdirectory. 
Just to mention a few of those example worlds, 
the warehouse demonstration file with a differential-driven robot is illustrated in  Figure~\ref{fig:screenshot.warehouse}, 
a greenhouse with a custom Ackermann-like mobile robot is shown in Figure~\ref{fig:screenshot.demos}(a), 
and a demonstration of how to run dozens of robots driving autonomously from Python scripts is shown in Figure~\ref{fig:screenshot.demos}(b).

\section{Impact}
\label{}


The present simulator has great potential to 
research on SLAM with either 
2D LiDAR \cite{grisetti2007improved,kohlbrecher2011flexible},
3D LiDAR \cite{hess2016real},
depth cameras \cite{whelan2016elasticfusion,mur2017orb}, 
or pure visual SLAM \cite{campos2021orb}, 
as well as collaborative SLAM approaches \cite{zou2019collaborative}
by means of multiple simultaneous robots. Generation of synthetic datasets with precise ground truth trajectories are key to train, test, and validate SLAM solutions.
MVSim can be also integrated with the ROS2 Nav2 \cite{macenski2022ros2} autonomous navigation framework, allowing easy debugging and testing
of navigation algorithms in simulation before moving them to real robots. 
Additionally, it allows ground-truth data generation for research
on vehicle dynamics and related topics, where real-world ground truth data is hard to obtain due to the nature of wheel-ground interactions \cite{reina2017vehicle,leanza2021factor}.





\section{Conclusions}
\label{}

After analyzing the design and features of the presented 
simulator, we defend that it has potential to
become one of the new tools researchers from different fields
might use in their day-to-day work to ease the generation of 
synthetic datasets, or test and validate new algorithms that 
require closing a loop between acting and sensing on moving agents.
In comparison to existing simulators, writing an MVSim XML world 
file to define a world and a vehicle or robot requires much less 
training and effort, lowering the entry barrier to ROS-based
simulation to students and researchers while still obtaining quality simulations.

\section{Conflict of Interest}

Authors declare that they have no competing financial interests 
or personal relationships that could have appeared 
to influence the work reported in this paper.

\section*{Acknowledgements}
\label{}

This work was partly supported by the Google Open Source Programs Office (OSPO) by 
means of a Google Summer of Code (GSOC) 2017 grant to Borys Tymchenko.
It has been also partially funded by the ``Agricobiot I" Project, UAL2020-TEP-A1991,
supported by the University of Almeria, Andalusian Economic Transformation, Industry,
Knowledge and Universities and ERDF funds, and ``Agricobiot II" Project, PY20\_00767,
supported by Andalusian Economic Transformation, Industry, Knowledge and Universities and ERDF funds.



\bibliographystyle{elsarticle-num} 
\bibliography{mvsim}

\begin{thebibliography}{10}
\expandafter\ifx\csname url\endcsname\relax
  \def\url#1{\texttt{#1}}\fi
\expandafter\ifx\csname urlprefix\endcsname\relax\def\urlprefix{URL }\fi
\expandafter\ifx\csname href\endcsname\relax
  \def\href#1#2{#2} \def\path#1{#1}\fi

\bibitem{grisetti2010tutorial}
G.~Grisetti, R.~K{\"u}mmerle, C.~Stachniss, W.~Burgard, A tutorial on
  graph-based slam, IEEE Intelligent Transportation Systems Magazine 2~(4)
  (2010) 31--43.

\bibitem{cadena2016past}
C.~Cadena, L.~Carlone, H.~Carrillo, Y.~Latif, D.~Scaramuzza, J.~Neira, I.~Reid,
  J.~J. Leonard, Past, present, and future of simultaneous localization and
  mapping: Toward the robust-perception age, IEEE Transactions on robotics
  32~(6) (2016) 1309--1332.

\bibitem{kavraki2016motion}
L.~E. Kavraki, S.~M. LaValle, Motion planning, in: Springer handbook of
  robotics, Springer, 2016, pp. 139--162.

\bibitem{mandow2007experimental}
A.~Mandow, J.~L. Martinez, J.~Morales, J.~L. Blanco, A.~Garcia-Cerezo,
  J.~Gonzalez, Experimental kinematics for wheeled skid-steer mobile robots,
  in: 2007 IEEE/RSJ international conference on intelligent robots and systems,
  IEEE, 2007, pp. 1222--1227.

\bibitem{reina2017vehicle}
G.~Reina, M.~Paiano, J.-L. Blanco-Claraco, Vehicle parameter estimation using a
  model-based estimator, Mechanical Systems and Signal Processing 87 (2017)
  227--241.

\bibitem{wang2016robust}
R.~Wang, H.~Jing, C.~Hu, M.~Chadli, F.~Yan, Robust h$\infty$ output-feedback
  yaw control for in-wheel motor driven electric vehicles with differential
  steering, Neurocomputing 173 (2016) 676--684.

\bibitem{meeussen2010rep}
W.~Meeussen, Rep 105-coordinate frames for mobile platforms, Available in:
  https://www.ros.org/reps/rep-0105.html (2010).

\bibitem{rico2022concise}
F.~M. Rico, A Concise Introduction to Robot Programming with ROS2, CRC Press,
  2022.

\bibitem{gerkey2003player}
B.~Gerkey, R.~T. Vaughan, A.~Howard, et~al., The player/stage project: Tools
  for multi-robot and distributed sensor systems, in: Proceedings of the 11th
  international conference on advanced robotics, Vol.~1, Citeseer, 2003, pp.
  317--323.

\bibitem{koenig2004design}
N.~Koenig, A.~Howard, Design and use paradigms for gazebo, an open-source
  multi-robot simulator, in: 2004 IEEE/RSJ International Conference on
  Intelligent Robots and Systems (IROS)(IEEE Cat. No. 04CH37566), Vol.~3, IEEE,
  2004, pp. 2149--2154.

\bibitem{Michel2004}
O.~Michel, \href{http://arxiv.org/abs/cs/0412052}{{WebotsTM: Professional
  Mobile Robot Simulation}} 1~(1) (2004) 39--42.
\newblock \href {http://arxiv.org/abs/0412052} {\path{arXiv:0412052}}.
\newline\urlprefix\url{http://arxiv.org/abs/cs/0412052}

\bibitem{hintjens2013zeromq}
P.~Hintjens, ZeroMQ: messaging for many applications, " O'Reilly Media, Inc.",
  2013.

\bibitem{catto2005iterative}
E.~Catto, Iterative dynamics with temporal coherence, in: Game developer
  conference, Vol.~2, 2005.

\bibitem{sola2018micro}
J.~Sola, J.~Deray, D.~Atchuthan, A micro lie theory for state estimation in
  robotics, arXiv preprint arXiv:1812.01537 (2018).

\bibitem{blanco2021tutorial}
J.~L. Blanco-Claraco, A tutorial on se(3) transformation parameterizations and
  on-manifold optimization, arXiv preprint arXiv:2103.15980 (2021).

\bibitem{ward2008dynamic}
C.~C. Ward, K.~Iagnemma, A dynamic-model-based wheel slip detector for mobile
  robots on outdoor terrain, IEEE Transactions on Robotics 24~(4) (2008)
  821--831.

\bibitem{horn1987closed}
B.~K. Horn, Closed-form solution of absolute orientation using unit
  quaternions, Journal of the Optical Society of America 4~(4) (1987) 629--642.

\bibitem{grisetti2007improved}
G.~Grisetti, C.~Stachniss, W.~Burgard, Improved techniques for grid mapping
  with rao-blackwellized particle filters, IEEE transactions on Robotics 23~(1)
  (2007) 34--46.

\bibitem{kohlbrecher2011flexible}
S.~Kohlbrecher, O.~Von~Stryk, J.~Meyer, U.~Klingauf, A flexible and scalable
  slam system with full 3d motion estimation, in: 2011 IEEE international
  symposium on safety, security, and rescue robotics, IEEE, 2011, pp. 155--160.

\bibitem{hess2016real}
W.~Hess, D.~Kohler, H.~Rapp, D.~Andor, Real-time loop closure in 2d lidar slam,
  in: 2016 IEEE international conference on robotics and automation (ICRA),
  IEEE, 2016, pp. 1271--1278.

\bibitem{whelan2016elasticfusion}
T.~Whelan, R.~F. Salas-Moreno, B.~Glocker, A.~J. Davison, S.~Leutenegger,
  Elasticfusion: Real-time dense slam and light source estimation, The
  International Journal of Robotics Research 35~(14) (2016) 1697--1716.

\bibitem{mur2017orb}
R.~Mur-Artal, J.~D. Tard{\'o}s, Orb-slam2: An open-source slam system for
  monocular, stereo, and rgb-d cameras, IEEE transactions on robotics 33~(5)
  (2017) 1255--1262.

\bibitem{campos2021orb}
C.~Campos, R.~Elvira, J.~J.~G. Rodr{\'\i}guez, J.~M. Montiel, J.~D. Tard{\'o}s,
  Orb-slam3: An accurate open-source library for visual, visual--inertial, and
  multimap slam, IEEE Transactions on Robotics 37~(6) (2021) 1874--1890.

\bibitem{zou2019collaborative}
D.~Zou, P.~Tan, W.~Yu, Collaborative visual slam for multiple agents: A brief
  survey, Virtual Reality \& Intelligent Hardware 1~(5) (2019) 461--482.

\bibitem{macenski2022ros2}
S.~Macenski, T.~Foote, B.~Gerkey, C.~Lalancette, W.~Woodall, Robot operating
  system 2: Design, architecture, and uses in the wild, Science Robotics 7~(66)
  (2022) eabm6074.

\bibitem{leanza2021factor}
A.~Leanza, G.~Reina, J.-L. Blanco-Claraco, A factor-graph-based approach to
  vehicle sideslip angle estimation, Sensors 21~(16) (2021) 5409.

\end{thebibliography}


\end{document}